# SCaRL- A Synthetic Multi-Modal Dataset for Autonomous Driving

Avinash Nittur Ramesh, Aitor Correas-Serrano, María González-Huici
Fraunhofer FHR, Germany

## Abstract

We present a novel synthetically generated multi-modal dataset, SCaRL, to enable the training and validation of autonomous driving solutions. Multi-modal datasets are essential to attain the robustness and high accuracy required by autonomous systems in applications such as autonomous driving. As deep learning-based solutions are becoming more prevalent for object detection, classification, and tracking tasks, there is great demand for datasets combining camera, lidar, and radar sensors. Existing real/synthetic datasets for autonomous driving lack synchronized data collection from a complete sensor suite. **SCaRL** provides synchronized **S**ynthetic data from RGB, semantic/instance, and depth *Cameras*; Range-Doppler-Azimuth/Elevation maps and raw data from ***R***adar; and 3D point clouds/2D maps of semantic, depth and Doppler data from coherent *Lidar*. SCaRL is a large dataset based on the CARLA Simulator, which provides data for diverse, dynamic scenarios and traffic conditions. SCaRL is the first dataset to include synthetic synchronized data from coherent Lidar and MIMO radar sensors.

## 1 Introduction

High-level environmental perception is essential for autonomous navigation in complex scenarios. Advanced driving assistance and autonomous navigation systems employ a combination of complementary and redundant sensors to perceive the surroundings and facilitate accurate decision-making. Currently, challenges in autonomous driving, such as obstacle detection, classification, and tracking, are being addressed through deep neural networks by training them on annotated datasets [1][2]. To further enhance and compare the performance of different neural networks and algorithms, choosing a dataset that offers sensor data from a complete sensor suite plays a vital role. Autonomous systems are equipped with several cameras, radars, and lidars to collect complementary information about their surroundings. Datasets created from such rigs require additional calibration, synchronization, and rectification as a post-processing task, which can be cumbersome and error-prone due to inherent limitations of the sensors and their placement, mismatch in the scanning rate and dimensionality of different sensors, environmental effects, etc. Moreover, measurement campaigns tend to capture sensor data in specific environments and traffic conditions with specific sensor systems, limiting the diversity in the dataset and promoting overfitting deep learning models. Photorealistic synthetic scenario simulation is gaining attention due to the possibility of including diverse and customizable scenarios in datasets for training or validating autonomous driving functionalities [3][4]. Acquiring a comparably diverse dataset from measurements would be expensive and potentially dangerous.

### 1.1 Related work

SYNTHIA [5], GTA V [6], Virtual KITTI [7], and Semantic KITTI [8] are some popularly used synthetic datasets for autonomous driving. SYNTHIA offers photorealistic RGB, depth, and semantic images. GTA V and Virtual KITTI datasets offer semantic/instance and RGB images. Semantic KITTI offers LiDAR point clouds with dense annotations. Simulators for FMCW radar in automotive scenarios can also be found in the literature: Wengerter et al. [9] presented an efficient FMCW radar simulator for multi-target traffic scenarios and showed the performance of a neural network based on EfficientDet for road user detection on Range-Doppler images. They proposed point target models for vulnerable road users, bicyclists, pedestrians, and vehicles and modeled their reflection characteristics and RCS. However, their scenarios are not dense enough to generate data for complex scenes (buildings, trees, etc.), and they use a single stationary radar. Schüßler et al. [10] proposed a radar ray tracing simulator based on material models for a complex scene modeled from the open-source autonomous driving simulator CARLA [11]. They address the problem of simulating multiple antennas and show that their simulator can reproduce measurement data realistically by comparing it with the range-azimuth maps. On the other hand, Dallmann et al. [12] simulated traffic scenarios using point scatter models for each road user entity defined in the scenario. They show that their approach is promising for testing automotive radar systems. However, both [10] and [12] investigate static scenarios and do not consider radar Doppler information or synchronized data from additional sensors.

### 1.2 Our contributions

None of the existing datasets provide a complete sensor suite with cameras, lidars, and radars for complex dynamic scenarios. To address the limitations mentioned in the previous section, we have generated the SCaRL dataset. SCaRL is a multi-modal dataset consisting of synthetic data from synchronized sensors such as RGB, semantic/instance, and depth cameras; radar raw data and Range-Doppler-Azimuth/Elevation maps; and coherent Lidar 3D point clouds and 2D semantic depth-Doppler maps.

In contrast to other datasets, we use CARLA to gather data from dynamic and complex pseudo-real-world scenarios involving various road users, from vehicles to pedestrians. We provide synchronized sensor data covering multiple viewpoints by enabling customizable sensor placement at arbitrary locations on a pseudo-real ego vehicle.

Synthetic data from virtual worlds such as SCaRL are subject to certain assumptions. Specifically, additive white Gaussian noise (AWGN) and linear motion models are considered. Thus, employing SCaRL to train neural networks to perform inference on real-world datasets requires an additional task of domain adaptation [13]. The benefits of using our large dataset are twofold: 1) reduced need for extensive and diverse real-world annotated datasets, and 2) faster convergence of training loss on a target real-world dataset by performing an initial training with our dataset, followed by transfer learning/domain adaptation [14].

## 2 Generation of SCaRL Dataset

The SCaRL dataset is primarily based on the CARLA simulator. CARLA is an open-source simulator developed on top of the Unreal game engine to support the development, training, and validation of autonomous urban driving scenarios. CARLA provides open digital assets (urban layouts, buildings, vehicles, pedestrians, street signs) that can be used freely. It is a simulation platform that allows the development of custom sensors, vehicles, and map layouts and provides support for enabling various environmental conditions. The assets, dynamic actors, maps, components, scenarios, and a few sensors present in CARLA are used as-is while generating our dataset. The steps involved in generating the dataset are as follows: 1) Software implementation of necessary code-base to enable extraction of additional details about all the actors in a scenario by diving deep into the basics of Unreal game engine and CARLA framework, 2) customization of the placement of sensors on the vehicle, 3) synchronization of sensor data, 4) transformation of raw sensor data into a representation favorable for fusion, and 5) addition of coherent Lidar and MIMO FMCW Radar to the simulation. By adding these sensors to the sensor suite of CARLA and processing acquired sensor data with our software, we provide a synchronized dataset containing the most diverse sensor suite publicly available.

### 2.1 Methodology

The main challenges in simulating coherent Lidar and MIMO FMCW radar sensors are related to the computational complexity and the need for a discrete representation of signals within the purview of the simulation platform. The physical model most often used for lidar and radar simulations is ray casting [10]. For the generation of the SCaRL dataset, we achieve this by modifying the existing lidar sensor of the CARLA simulator to obtain additional information from the Unreal game engine, such as angle of incidence, type of target, absolute 3D position and speed, and surface orientation. We obtain a dense point cloud with rich environmental information by casting multiple such rays to encompass the scenario in both azimuth and elevation. This is achieved in a post-processing stage that transforms rich 3D lidar point clouds into the desired data: 2D semantic, depth, Doppler maps for lidar; and raw data and Range-Doppler-Azimuth/elevation maps for radar.

### 2.2 Simulation of Coherent LiDAR

Coherent lidars are a new generation based on FMCW optical carriers [15]. They promise to reach long distances, have built-in interference mitigation from sunlight, and can measure Doppler. To simulate this lidar, we use the rotating lidar implemented in CARLA. These sensors obtain depth and semantic information from the casted ray at the point of incidence on the target. We modify the code base to calculate the target's relative velocity, the ray's reflection intensity based on surface normal, angle of incidence, and distance-based non-linear attenuation. Furthermore, we calculate the intrinsic camera calibration matrix to obtain 2D real-lidar-like semantic, depth, and Doppler maps and transform the point cloud from world coordinates to camera coordinates. We randomly annihilate specific pixels based on reflection intensity and introduce AWGN with variance proportional to the distance. The result is a Doppler-aware lidar simulation with enhanced physics simulation by adding material, orientation, and dynamic characteristics of the targets and ego-vehicle.

### 2.3 Simulation of MIMO FMCW Radar

Consider a MIMO FMCW radar with $N_{Tx}$ transmit antennas and $N_{Rx}$ receive antennas. The $m$-th transmitted chirp in each burst is

$$s(t,m) = A_{Tx} \exp\left(2\pi(f_0 t + \frac{k}{2}t^2)\right) \quad (1)$$

Where $k$ is the chirp modulation rate and $f_0$ is the carrier frequency (e.g., 77 GHz for automotive radar). For target $t$ and after dechirping, the received signal in the beat-frequency domain can be approximated as

$$r_t(t,m) = A_{Tx_t} A_{Rx_t} \exp\left(j2\pi(f_{b_t}t + f_{d_t}mT + f_c\tau_0)\right) \quad (2)$$

Where $f_b = k\tau_0$ is the beat frequency, $f_d = 2v_0 f_c c^{-1}$ is the Doppler frequency, and $T$ is the time between chirps. For $N_t$ targets, the received signal is

$$r(t,m) = \sum_{t=1}^{N_t} r_t(t,m) \quad (3)$$

Note that the amplitudes $A_{Tx}$ and $A_{Rx}$ are dependent on the transmitted and received power $P_{Tx}$ and $P_{Rx}$ respectively. While $P_{Tx}$ is a simulation parameter, $P_{Rx}$ is calculated through the radar equation

$$P_{RX} = \frac{P_{TX}\, G_{TX}(\alpha,\gamma) G_{RX}(\alpha,\gamma)\, \sigma\, \lambda^2}{(4\pi)^3\, d^4}, \quad \lambda = \frac{f_0}{c} \quad (4)$$

which models the angle-dependent transmit and receive antenna gains $G_{TX}(\alpha,\gamma)$ and $G_{RX}(\alpha,\gamma)$, as well as the target radar-cross-section $\sigma$ and the free-space loss. The targets' scattering patterns are assumed to be isotropic, and Fresnel's law of reflection is used to scale down reflected power. Other effects, such as probabilistic weather attenuation, can be easily included in (5) to model diverse atmospheric conditions.

Table 1. Design parameters for MIMO FMCW Radar

| Parameters | Value (SI Units) |
|---|---|
| Bandwidth, B | $300\ MHz$ |
| Chirp time, $T_c$ | $30\ \mu s$ |
| Inter-chirp time, $T_i$ | $8\mu s$ |
| No. of chirps per frame, $N_d$ | 256 |
| No. of samples per chirps, $N_r$ | 512 |
| Azimuth resolution | $9.549°$ |
| Azimuth FoV (HPBW) | $\pm 45°$ |

The model above considers a single transmitter and receiver. To simulate MIMO paths, the distance between each transmit and receive antenna is calculated through ray-casting. The path difference between channels translates into different $\tau_0$, which in turn causes a phase difference amongst MIMO channels that is captured in the term $\exp(j2\pi f_c \tau_0)$ in (2). This captures a near-field model that is robust to the usage of very large arrays and the presence of very close targets.

The typical DFT-based processing is used within the simulator for range, Doppler, and angle estimation. First, the received signal for each MIMO channel is discretized and rearranged into a data matrix $\mathbf{D} \in \mathbb{C}^{N_r \times N_d}$, where $N$ is the number of fast-time samples, and $M$ is the number of chirps in a sequence. The range-Doppler map is extracted using a 2D-FFT of $\mathbf{D}$. To estimate the direction of arrival (DoA) in each cell of the range-Doppler map, the entries for each MIMO channel are aligned to form the measurement vectors $\boldsymbol{a}_{r,d} \in \mathbb{C}^{1 \times N_{MIMO}}$, where $N_{\text{MIMO}}$ is the number of MIMO channels (or elements in the virtual array). If a uniform virtual array is used and far-field is assumed, the angular spectrum of each cell can be calculated with another DFT. This results in the range-Doppler-azimuth/elevation 3D map, of which any 2D cuts can be visualized. The MIMO channel simulation of SCaRL is general enough that sparse MIMO front-ends can be considered for enhanced resolution; in this case, different DoA estimation approaches, such as the ones presented in [16], can be used. Moreover, the simulation of complex sensor suites such as the vehicle-mounted sparse radar networks studied in [17] is also compatible with SCaRL.

### 2.4 Synthetic scenarios

The SCaRL dataset consists of 140000 time synchronized and aligned data frames from 6x RGB, 6x semantic/instance segmentation, 6x depth cameras, 6x lidar, and 6x MIMO FMCW radar sensors. Each sensor suite is spaced $0.4m$ apart along a horizontal. The sensors are placed in the front and back of the ego-vehicle, as shown in Figure 1, and are driven in all the seven scenarios provided by CARLA. Radar is simulated by illuminating the scene with $10^6$ rays in a fixed grid pattern, resulting in a dense point cloud (considered as point targets) with rich environmental information (such as angle of incidence and surface orientation to determine the reflected power, target class to determine the RCS, position, velocity, etc.).

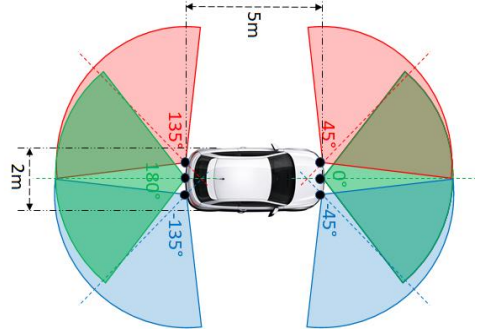

Figure 1. Placement of complete sensor suites (containing five sensors) at six positions (black dot) around the ego vehicle

To showcase radar scenario simulation, we replicate the Texas Instruments AWR1843BOOST [18], with its waveform parameters described in Table 1. Figure 2 shows an example of the SCaRL dataset when this radar is considered. Although the angular resolution in the simulated radar is relatively low, a few key features in the simulated scene can be observed in the radar estimation. Most prominently, the guardrail can be seen as an extended target spanning from -10 to 50 degrees at varying ranges and negative Doppler (as the moving platform is approaching it). Other static targets can be seen within the same Doppler band, and the car in front can be distinguished in the range-Doppler map as the only target with positive velocity. This simple simulation shows that extended targets are correctly simulated as radar data and synchronized with both camera and coherent Lidar data.

## 3 Summary and Future Work

This paper has presented the SCaRL dataset, a multi-modal synthetic dataset that provides synchronized photorealistic measurements from camera, (coherent) lidar, and radar. The generation of synthetic range-Doppler-angular data of arbitrarily placed FMCW MIMO automotive radars in dynamic automotive scenarios is presented to showcase the potential of SCaRL. SCaRL is intended to assist in the training and validation of neural networks and in comparing different waveforms, sensors, and estimation approaches. SCaRL provides data from a complete sensor suite and encourages the development of sensor fusion algorithms. As the radar simulation is built upon a ray-casting model, it is possible to implement other MIMO waveforms using the same framework. Extensions to PMCW, PC-FMCW, and OFDM waveforms are planned to evaluate next-generation radar waveforms in realistic scenarios. Moreover, future development will also be oriented towards further reducing the gap between simulation and reality by considering the effects of target material, multi-path propagation, free-path loss, and different array configurations.

## 4 References

[1] A J. Martin, D.Vazquez, D. Geronimo, and A.Lopez. Learning appearance in virtual scenarios for pedestrian detection. In IEEE (CVPR), 2010.

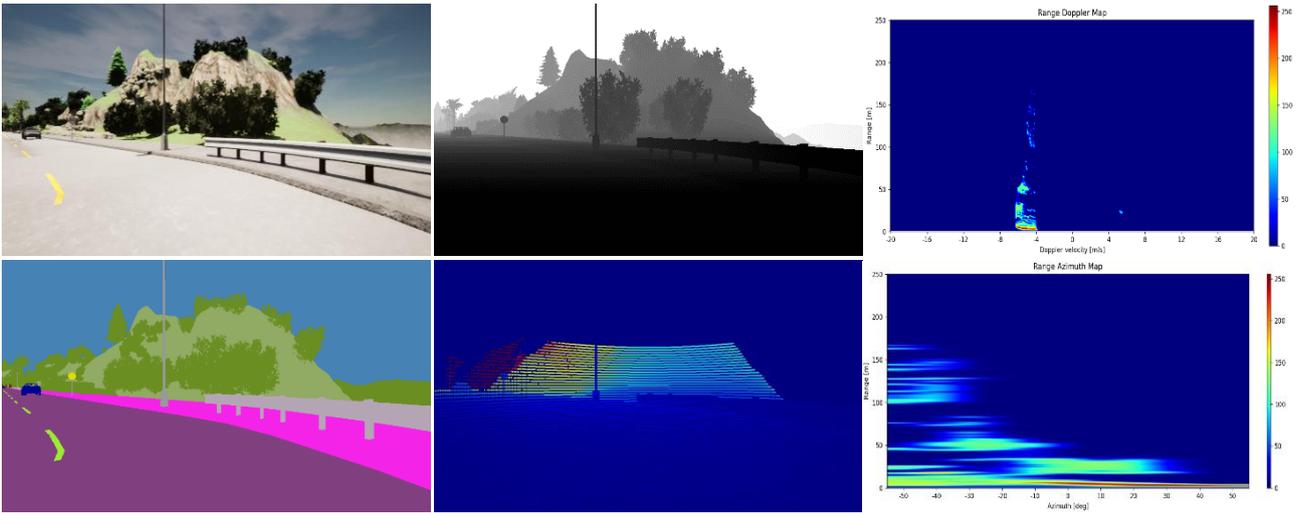

Figure 2: Scarl dataset with synchronized RGB, depth images and Range-Doppler map in the first row, and seman-tic/in-stance segmentation image, 2D LiDAR depth map and Range-Azimuth map for RADAR in the second row.

[2] C. J. Papon and M. Schoeler. Semantic pose using deep networks trained on synthetic RGB-D. In Intl. Conf. on Computer Vision (ICCV), 2015.

[3] C. Schüßler, M. Hoffmann, J. Bräunig, I. Ullmann, R. Ebelt and M. Vossiek, "A Realistic Radar Ray Tracing Simulator for Large MIMO-Arrays in Automotive Environments," in IEEE Journal of Microwaves, vol. 1, no. 4, pp. 962-974, Oct. 2021, doi: 10.1109/JMW.2021.3104722.

[4] F. Rutz, R. Rasshofer and E. Biebl, "Transfer Learning in Automotive Radar Using Simulated Training Data Sets," 2023 24th International Radar Symposium (IRS), Berlin, Germany, 2023, pp. 1-7, doi: 10.23919/IRS57608.2023.10172422.

[5] Ros, G., Sellart, L., Materzynska, J., Vazquez, D. and Lopez, A.M., 2016. The synthia dataset: A large collection of synthetic images for semantic segmentation of urban scenes. In Proceedings of the IEEE conference on CVPR (pp. 3234-3243).

[6] Richte, S.R., Vineet, V., Roth, S. and Koltun, V., 2016. Playing for data: Ground truth from computer games. In Computer Vision–ECCV 2016, The Netherlands, October 11-14, 2016, Proceedings, Part II 14 (pp. 102-118). Springer International Publishing.

[7] Gaido, A. et al., 2016. Virtual worlds as proxy for multi-object tracking analysis. In Proceedings of the IEEE CVPR (pp. 4340-4349).

[8] Behle, J. et al., 2019. Semantickitti: A dataset for semantic scene understanding understanding of lidar sequences. In Proceedings of the IEEE/CVF international conference on computer vision (pp. 9297-9307).

[9] Wengerter, T., Pérez, R., Biebl, E., Worms, J. and O'Hagan, D., 2022, June. Simulation of urban automotive radar measurements for deep learning target detection. In 2022 IEEE Intelligent Vehicles Symposium (IV) (pp. 309-314). IEEE.

[10] Schüßler, C., Hoffmann, M., Bräunig, J., Ullmann, I., Ebelt, R. and Vossiek, M., 2021. A realistic radar ray tracing simulator for large MIMO-arrays in automotive environments. IEEE Journal of Microwaves, 1(4), pp.962-974.

[11] Dosovitskiy, A., Ros, G., Codevilla, F., Lopez, A. and Koltun, V., 2017, October. CARLA: An open urban driving simulator. In Conference on robot learning (pp. 1-16). PMLR.

[12] T. Dallmann, J. K. Mende, S. Wald. "ATRIUM: A Radar Target Simulator for Complex Traffic Scenarios," in proceedings of the International Conference on Microwaves for Intelligent Mobility (ICMIM), Munich, 2016, 1-4. 10.1109/ICMIM.2018.8443515.

[13] D. Vazquez, A. Lopez, J. Marın, D. Ponsa, and D. Geronimo. Virtual and real world adaptation for pedestrian detection. IEEE Trans. Pattern Anal. Machine Intell., 2014.

[14] Ramesh, A.N. et al, 2023. SIUNet: Sparsity Invariant U-Net for Edge-Aware Depth Completion. In Proceedings of the IEEE/CVF WACV (pp. 5818-5827).

[15] S. Cwalina et al., "Fiber-based Frequency Modulated LiDAR With MEMS Scanning Capability for Long ange Sensing in Automotive Applications," 2021 IEEE MetroAutomotive, Italy, 2021, pp. 48-53.

[16] Correas-Serrano, A., & González-Huici, M. A. (2018, June). Experimental evaluation of compressive sensing for doa estimation in automotive radar. In 2018 19th International Radar Symposium (IRS) (pp. 1-10). IEEE.

[17] Correas-Serrano, A., Gonzalez-Huici, M., Simoni, R., Bredderman, T., Warsitz, E., Müller, T., & Kirsch, O. (2022, September). Performance Analysis and Design of a Distributed Radar Network for Automotive Application. In *2022 23rd International Radar Symposium (IRS)* (pp. 30-35). IEEE.

[18] AWR1843AOP Single-chip 77- and 79-GHz FMCW mmWave Sensor AOP datasheet (Rev.B).